%% file: portuguese-tiles.tex
\documentclass[sigconf]{acmart}

\usepackage{booktabs} 

\usepackage{times}
\usepackage{epsfig}
\usepackage{graphicx}
\usepackage{amsmath}
\usepackage{amssymb}
\usepackage{multirow}
\usepackage{flushend}
\usepackage{subcaption}
\usepackage{gensymb}
\usepackage{algorithm}
\usepackage[noend]{algpseudocode}
\graphicspath{{figures/}}

\newcommand{\etal}{\textit{et al.}}
\newcommand{\ie}{\textit{i.e.}}
\newcommand{\eg}{\textit{e.g.}}

\setcopyright{rightsretained}

\copyrightyear{2019}
\acmYear{2019}
\setcopyright{acmlicensed}
\acmConference[GECCO '19]{Genetic and Evolutionary Computation Conference}{July 13--17, 2019}{Prague, Czech Republic}
\acmBooktitle{Genetic and Evolutionary Computation Conference (GECCO '19),
July 13--17, 2019, Prague, Czech Republic}
\acmPrice{15.00}
\acmDOI{10.1145/3321707.3321821}
\acmISBN{978-1-4503-6111-8/19/07}

\usepackage[overlay,absolute]{textpos}

\begin{document}

\begin{textblock*}{10in}(22mm, 10mm)
{\textbf{Ref:} \emph{ACM Genetic and Evolutionary Computation Conference (GECCO)}, pages 1319-1327, Prague, Czech Republic, July 2019.}
\end{textblock*}

\title{A Novel Hybrid Scheme 
Using Genetic Algorithms and Deep\\
Learning for the Reconstruction of Portuguese Tile Panels}

\author{Daniel Rika}
\affiliation{%
   \institution{Dept.~of Computer Science\\Bar-Ilan University}
   \city{Ramat-Gan 52900} 
   \state{Israel} 
}
\email{danielrika@gmail.com}

\author{Dror Sholomon}
\affiliation{%
   \institution{Dept.~of Computer Science\\Bar-Ilan University}
   \city{Ramat-Gan 52900} 
   \state{Israel} 
}
\email{dror.sholomon@gmail.com}

\author{Eli (Omid) David}
\affiliation{%
   \institution{Dept.~of Computer Science\\Bar-Ilan University}
   \city{Ramat-Gan 52900} 
   \state{Israel} 
}
\email{mail@elidavid.com}

\author{Nathan S.~Netanyahu}
\authornote{Also affiliated with the Gonda Brain Research Center at Bar-Ilan University, and the Center for Automation Research, University of Maryland, College Park, MD 20742.}
\affiliation{%
   \institution{Dept.~of Computer Science\\Bar-Ilan University}
   \city{Ramat-Gan 52900} 
   \state{Israel} 
}
\email{nathan@cs.biu.ac.il}

\renewcommand{\shortauthors}{D. Rika, D. Sholomon., E.O. David, and N.S. Netanyahu}

\begin{abstract}
This paper presents a novel scheme, based on a unique combination of {\em genetic algorithms} (GAs) and {\em deep learning} (DL), for the automatic reconstruction of {\em Portuguese tile panels}, a challenging real-world variant of the {\em jigsaw puzzle problem} (JPP) with important national heritage implications. Specifically, we introduce an enhanced GA-based puzzle solver, whose integration with a novel DL-based {\em compatibility measure} (DLCM) yields state-of-the-art performance, regarding the above application. Current compatibility measures consider typically (the chromatic information of) edge pixels (between adjacent tiles), and help achieve high accuracy for the synthetic JPP variant. However, such measures exhibit rather poor performance when applied to the Portuguese tile panels, which are susceptible to various real-world effects, e.g., monochromatic panels, non-squared tiles, edge degradation, etc. To overcome such difficulties, we have developed a novel DLCM to extract high-level texture/color statistics from the entire tile information.

Integrating this measure with our enhanced GA-based puzzle solver, we have demonstrated, for the first time, how to deal most effectively with large-scale real-world problems, such as the Portuguese tile problem. Specifically, we have achieved 82\% accuracy for the reconstruction of Portuguese tile panels with unknown piece rotation and puzzle dimension (compared to merely 3.5\% average accuracy achieved by the best method known for solving this problem variant). The proposed method outperforms even human experts in several cases, correcting their mistakes in the manual tile assembly.
\end{abstract}

%
%



\maketitle

\input{DLCM_body_conf}

\bibliographystyle{ACM-Reference-Format}
\bibliography{portuguese-tiles} 

\end{document}

%% file: DLCM_body_conf.tex
\begin{figure}[h]
\centering
    \begin{subfigure}[t]{0.48\linewidth}
        \includegraphics[width=\linewidth]{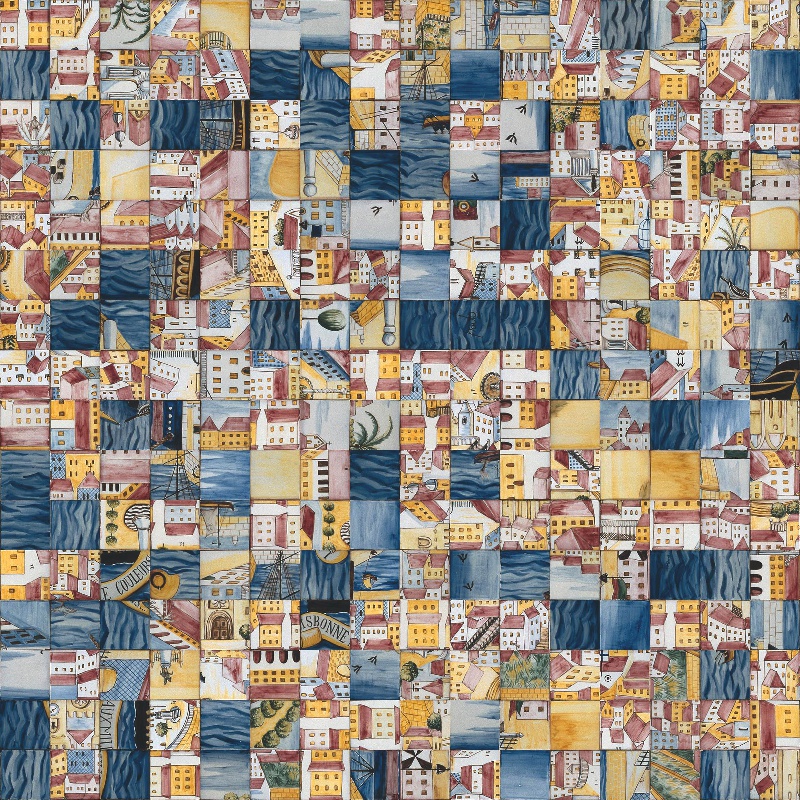}
    \end{subfigure}
    \begin{subfigure}[t]{0.48\linewidth}
        \includegraphics[width=\linewidth]{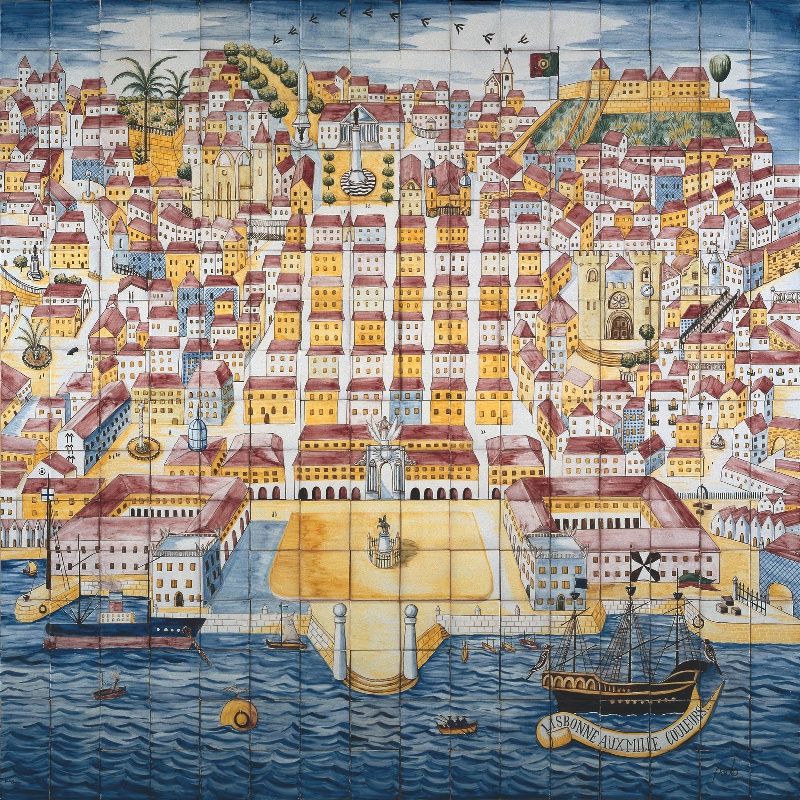}
    \end{subfigure}
    \par\medskip
    \begin{subfigure}[t]{0.48\linewidth}
        \includegraphics[width=\linewidth]{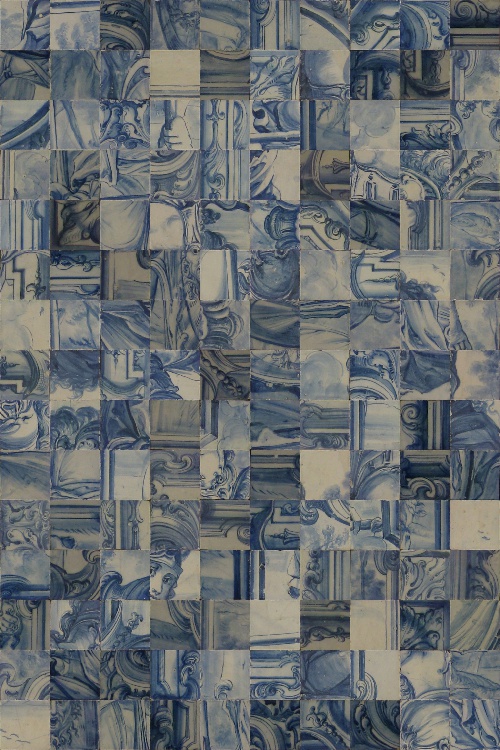}
    \end{subfigure}
    \begin{subfigure}[t]{0.48\linewidth}
        \includegraphics[width=\linewidth]{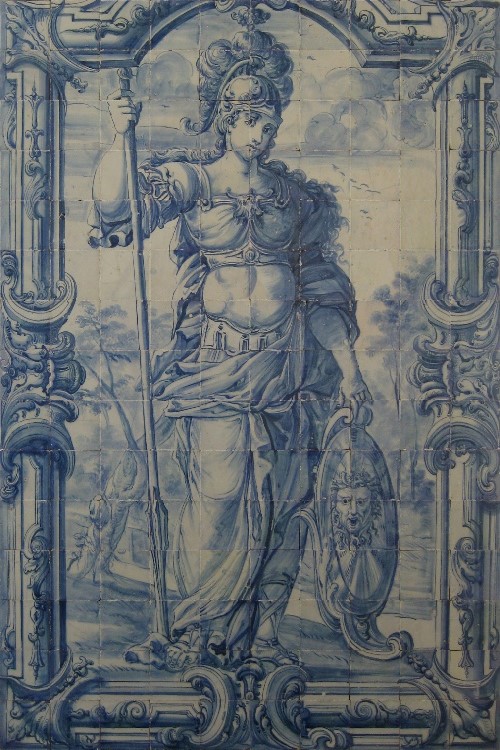}
    \end{subfigure}
\caption{Reconstruction of Portuguese tile panels with unknown piece orientation and panel dimensions, due to our proposed system. \textbf{Left:} Input images of Portuguese tile panels, containing 256 (top) and 150 (bottom) pieces. \textbf{Right:} Perfectly reconstructed images due to our novel compatibility measure coupled with the enhanced version of a "kernel-growth" GA.}
\label{fig:ga_reconstruction_samples}
\end{figure}

\section{Introduction}
    Object reconstruction from numerous fragments is a pervasive, important task that has been encountered in many areas throughout human civilization. Piecing together broken pottery, ancient frescoes, or shredded documents from their artifacts are merely a few examples. The most basic generic version of the problem is to assemble an object from its $n$ different (non-overlapping) pieces as accurately and efficiently as possible. (To automate this challenging task, processing is often applied to colored images acquired from these pieces.) The basic problem definition is very similar to the popular {\em jigsaw puzzle problem} (JPP), which is known to be NP 
    complete~\cite{journals/aai/Altman89,springerlink:10.1007/s00373-007-0713-4}. The JPP has been pursued by many researchers, as it is a special instance of a broad class of challenging real-world problems, such as image editing~\cite{bb43059}, the recovery of shredded documents or photographs~\cite{liu2011automated,marques2009reconstructing,justino2006reconstructing,conf/icip/DeeverG12}, art conservation~\cite{journals/tog/BrownTNBDVDRW08,journals/KollerL06,andalo2016impa}, speech descrambling~\cite{Zhao:2007:PSA:1348258.1348289,chuman2017}, etc., as well as additional problems in areas like 
    biology~\cite{journals/science/MarandeB07}, chemistry~\cite{oai:xtcat.oclc.org:OCLCNo/ocm45147791},
    literature~\cite{conf/ifip/MortonL68}, and more.
    Obviously, there are notable differences, in practice, between a pure JPP setting and the above real-world problems ({\eg} unknown dimensions, missing pieces, gaps between pieces due to degradation over time, pieces from multiple puzzles, etc.). Nevertheless, the JPP serves as a testbed for developing ground-breaking methods for these important challenges.
    
    Every reconstruction procedure requires a {\em compatibility measure} to estimate the likelihood that two given pieces are adjacent and a strategy for placing the pieces as ``accurately'' as possible with respect to some global objective function.
    Although much effort has been devoted to devising reliable compatibility measures for jigsaw-like problems, they may not always be consistent\footnote{In the sense that the most compatible piece to a given piece $A$, with respect to a compatibility measure in question, may not necessarily be adjacent to $A$ in the ``correct'' puzzle configuration.}; if they were, the problem would not be NP-hard. More importantly, the typical dependence of current compatibility measures on correlations between low-level color/texture statistics in the proximity of tile boundaries, renders jigsaw puzzle solvers based on such measures virtually ineffective for real-world problems, such as the reconstruction of 
    archaeological fragments and shredded documents (where often the information is severely degraded near the points of fraction), or that of Portuguese tile panels, whose image content is not necessarily color-rich and where chromatic information near tile boundaries might be severely corrupted. In addition, many methods for solving optimally the piece placement problem resort to greedy strategies, which are problematic in encountering local optima. Moreover, they usually cannot recover from erroneous placements made early on (as a result of a greedy, locally optimal choice). To meet these challenges, we employ in this paper a {\em computational intelligence} (CI) approach in dealing effectively with {\em both} components of the problem ({\ie} the search and the compatibility measure).
    Specifically, we present a {\bf unique combination} of: (1) An enhanced {\em genetic algorithm} (GA)-based scheme
    for finding promising (partial) solutions ({\ie} {\em fittest chromosomes}), at each iterative stage, as a strategy for optimal piece placement, and (2) a novel {\em deep learning} (DL) model for learning piece compatibility by directly training on the raw data (of a fairly small training set), without applying any standard feature selection/extraction techniques,
    
    Our contributions are summarized as follows:
    \begin{enumerate}
    \item
    Provided an enhanced GA solver for the construction of Portuguese tile panels;
    \item
    Obtained for the first time a DL-based compatibility measure (DLCM) for a real-world JPP-like task;
    \item
    Presented a unique combination of the above GA module and the novel compatibility measure for the reconstruction of Portuguese tile panels on a large-scale basis (see {\eg} Fig.~\ref{fig:ga_reconstruction_samples});
    \item
    Obtained state-of-the-art-results for the above real world problem; specifically, achieved an average accuracy of 82\% on Type 2 puzzles with unknown dimensions (compared to merely 3.5\% average accuracy achieved by Gallagher's method~\cite{conf/cvpr/Gallagher12}, which is the best method known for solving this problem variant);
    \item
    Compiled a new benchmark for the community, regarding training and test data for the Portuguese tile problem. \end{enumerate}
    
    The paper is organized as follows. Section~\ref{RelWrk} provides a brief survey of recent related work. Section~\ref{sec:ga_solver} and Section~\ref{sec:dl_cm} describe, respectively, our novel GA-based solver and the DL method for learning a compatibility measure. Section~\ref{sec:datasets} presents the datasets used, and Section~\ref{sec:experimental_results} provides detailed experimental results. Section~\ref{sec:conclusions} makes concluding remarks.

\section{Related Work}\label{RelWrk}
    \subsection{Synthetic JPP}
        \subsubsection{Traditional Methods}
            Freeman and Garder~\cite{bb47278} introduced initially in 1964 a computational solver, which handled up to nine-piece puzzles. Subsequent research~\cite{radack1982jigsaw,wolfson1988solving,kong2001solving,goldberg2002global} relied solely on shape cues of the pieces.
            Kosiba {\etal}~\cite{kosiba1994automatic} were the first to use image content, in addition to boundary shape; their method computes color compatibility along the matching contour, rewarding adjacent jigsaw pieces with similar colors. This trend continued for more than a decade (see, {\eg}  \cite{chung1998jigsaw,yao2003shape,makridis2006new,sagiroglu2006texture,nielsen2008solving}), before the research focus shifted from shape-based to merely color-based solvers of square-tile puzzles with known piece orientation ({\ie} Type 1 puzzles).
            
            Cho {\etal}~\cite{conf/cvpr/ChoAF10} used {\em dissimilarity} ({\ie} the sum, over all neighboring pixels, of squared color differences over all color bands), as a compatibility measure for their probabilistic puzzle solver, that handles up to 432 pieces, given some a priori knowledge of the puzzle. (The sum of squared differences is referred to as SSD.) Their 2010 paper was followed by Yang {\etal}~\cite{yang2011particle}, who reported improved performance due to their {\em particle filter}-based solver. Shortly after, Pomeranz {\etal}~\cite{conf/cvpr/PomeranzSB11} presented, for the first time, a fully-automated jigsaw puzzle solver of puzzles containing up to 3,000 square pieces, using the above defined dissimilarity and their so-called {\em best-buddies} heuristic.  
            Gallagher~\cite{conf/cvpr/Gallagher12} advanced further the state-of-the-art by considering a more general variant of the problem, where a piece orientation is unknown ({\ie} Type 2 puzzle), as well as the puzzle dimensions. Specifically, he presented the preferable measure of {\em Mahalanobis gradient compatibility} (MGC), which penalizes changes in intensity gradients (rather than changes in intensity) and learns the covariance of the color channels, using the Mahalanobis distance. He suggested also {\em dissimilarity ratios} for a more indicative compatibility measure.
            
            Sholomon {\etal}~\cite{Sholomon_2013_CVPR,sholomon2014genetic,sholomon2014generalized} pursued a GA-based approach based on a number of innovative {\em crossover} procedures, and demonstrated the effective performance of their methodology on very large Type 1 and Type 2 puzzles (including two-sided puzzles and a number of mixed puzzles). Son \textit{et al.}~\cite{son2014solving} imposed so-called {\em loop constraints}, where the dissimilarity ratio (with respect to the smallest distance from a piece edge in question), for each consecutive pair of pieces along a loop of four or more pieces, is below a certain threshold. They were able to improve the accuracy for both Type 1 and Type 2 puzzles in certain cases. Also, they provided, for the first time, an upper bound on the reconstruction accuracy for various datasets. Paikin and Tal~\cite{paikin2015solving} proposed a greedy solver based on an asymmetric $L_1$-norm dissimilarity and the best-buddies heuristic. They demonstrated how to handle, among other things, puzzles with missing pieces, and reported improved accuracy results and fast running times. More recently, Andal{\'o} {\etal}~\cite{7442162} showed how to map the JPP to the problem of maximizing a constrained quadratic function, and presented a deterministic algorithm for solving it via gradient ascent.
        
        \vspace{-2pt}
        \subsubsection{DL Methods}
            Recently, there have been also a few DL works related to the JPP~\cite{doersch2015unsupervised,noroozi2016unsupervised,dery2017jpp,santacruz2017visualpermutation}. However, these works barely provide any practical solutions to even ``toy instances'' of the JPP, and their main thrust is to ``re-purpose'' a neural network, trained to solve a simple jigsaw puzzle (without manual labeling), to handle advanced tasks, such as object detection and classification, in an unsupervised manner. Other than the above, a DL-based heuristic called {\em DNN-buddies} was presented in~\cite{sholomon2016dnn}, in an attempt to enhance the accuracy of a GA-based solver. It should be noted, though, that the above heuristic is employed in conjunction with the SSD measure, in a rather restrictive manner, so it is expected to perform rather poorly on real-world JPP-like tasks.   
    
    \subsection{Real-World Portuguese Tile Panels}
        The reconstruction of ancient frescoes and wall paintings from numerous large repositories of fragmented artifacts, compiled over time due to natural deterioration, is of utmost importance in preserving world cultural heritage. Various efforts to automate the process ({\eg}\cite{ieeetsp2002,journals/tog/BrownTNBDVDRW08,eurographics2016}) rely primarily on shape matching (in 2D and 3D) of fragments followed by their assembly. While exhibiting good performance on relatively small datasets (only a few hundred fragments), the scalability of these efforts (in terms of the number of fragments and the number of art works in a given pool) is questionable.
        
\begin{figure}[h]
        \centering
        \includegraphics[width=\linewidth]{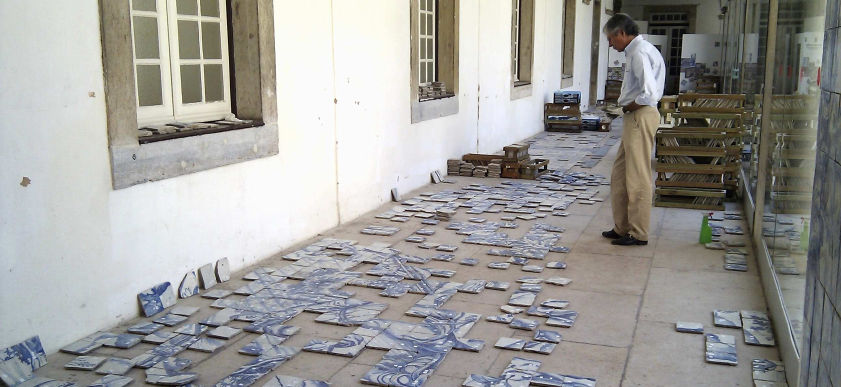}
        \caption{Manual assembling of a panel of Portuguese tiles at the National Tile Museum (Museu Nacional do Azulejo, MNAz), Lisbon, Portugal: Source~\cite{fonseca2012montagem}.}
        \label{2D_Tiles}
        \end{figure}

        Our focus in this paper is on the reconstruction of the Portuguese tiles panels~\cite{de2011azulejos}, which concerns the assembly of ancient panels of 2D square tiles that have been removed from many buildings and landmarks in Portugal (see Figure~\ref{2D_Tiles}). Currently, over one hundred thousand such tiles are stored at the Portuguese National Tile Museum (Museu Nacional do Azulejo) in Lisbon, and are awaiting manual assembly by human experts. In view of the extremely challenging nature of the problem, it would take decades, at the current pace, before all these ``jigsaw puzzles'' are solved, {\ie} before the panels are assembled by the human experts~\cite{pais18}.
        
        Fonseca~\cite{fonseca2012montagem} acquired tile images and adapted their shape to squares; he then applied an augmented Lagrange multipliers technique to an equivalent optimization problem and a greedy approach for Type 1 and Type 2 variants, respectively. He obtained 57.8\% and 39.1\% accuracy for these cases, respectively, on panels containing only a few dozen tiles.  In comparison, Gallagher's method~\cite{conf/cvpr/Gallagher12} achieves corresponding accuracy levels of 64.5\% and 49.4\%. Andalo {\etal}~\cite{andalo2016impa} reported perfect reconstruction (of 4 mixed tile panels) using their PSQP method~\cite{7442162} for known tile orientation. However, their method does not handle the Type 2 variant, and its preliminary results were obtained for panels containing a fairly small number of, presumably, high-resolution tiles.  
        
\section{GA Solver}
\label{sec:ga_solver}
    
    We seek a global optimizer that can exploit the relative accurate piece adjacency prediction capability, but that can also overcome its inaccuracies. Previous solvers rely typically on some specialized criterion, which implies a subset of edge adjacencies that are likely to be correct. To avoid searching for such a specific criterion, we pursue a GA approach~\cite{holland1975adaptation} for tile placement, in the spirit of the {\em kernel-growth} scheme presented in Sholomon  {\etal}~\cite{Sholomon_2013_CVPR, sholomon2014generalized}.
    Since the proposed GA solver is of a random nature, it could correct, potentially, wrong adjacencies during the global optimization.
    
    Following~\cite{Sholomon_2013_CVPR}, we describe here the new hierarchical phases of our modified {\em crossover} operator. In a nutshell, a chromosome is associated with a puzzle configuration (or a ``solution''), and its fitness function is defined by the overall sum of pairwise, adjacent tile compatibilities (see below). The principle of hierarchical phases is that a piece is added to the growing kernel at each phase only if the previous phases have been exhausted ({\ie} no further pieces can be added due to these phases); the crossover terminates once the kernel contains all the pieces. Our proposed phases and their hierarchical arrangement are as follows.
    
    \begin{itemize}
        \item \textbf{Phase I:} If there is a free (piece) boundary in the kernel, which has a neighboring piece in a chromosome parent, such that the score of each of these adjacent pieces is greater than ${\rm {max}}(0.8, C_{\rm{mean}})$, where $C_{\rm{mean}}$ is the chromosome's average compatibility across all boundaries, then add the neighboring piece to the kernel. We define the score of a piece as the average compatibility measure between the piece and all of its neighbors. This phase gives priority to the chromosome parent with the higher fitness, assuming that it would yield a more accurate reconstruction rate.
        
        \item \textbf{Phase II:} Similar to Phase I, except that this phase selects the chromosome parent with the lower fitness.
        
        \item \textbf{Phase III:} If there is a free (piece) boundary in the kernel, such that the two chromosome parents agree on the adjacent piece, place this piece next to the boundary in question.
        
        \item \textbf{Phase IV:} If there is a free (piece) boundary in the kernel, such that its most compatible piece is available ({\ie} is not placed already in the kernel), then add that compatible piece to the kernel.
        
        \item \textbf{Phase V:} If there is a free (piece) boundary in the kernel, such that its second-most compatible piece is available, then add the latter piece to the kernel.
        
        \item \textbf{Phase VI:} Pick randomly one of the remaining pieces, and place it randomly at one of the free boundaries of the kernel.
    \end{itemize}
    
    We introduce a certain degree of randomness to the process (known as {\em mutation}), in order to avoid local maxima, by skipping some of the crossover phases, with small probability.
    Specifically, we skip the first and second phases with 10\% probability and the third phase with 20\% probability. The other phases are always executed.
    
    Other hyper-parameters of our modified GA solver (which were arrived at after exhaustive experimentation) are as follows: Chromosomes are chosen for the crossover operation according to the {\em roulette wheel selection}, the population consists of 100 chromosomes, and the GA runs for 500 generations.
    
    \subsection{Rationale}
        Before explaining the rationale behind the above phases, we note that our proposed crossover does not draw on the notion of {\em best-buddies}, as was defined and used {\eg} in~\cite{conf/cvpr/PomeranzSB11} and~\cite{Sholomon_2013_CVPR}.
        The reason for that is that in contrast to the (synthetic) JPP, where best-buddy pairs were found to be adjacent with 95\% probability, our experience with the Portuguese tile panels shows that best-buddy pairs, with respect to our state-of-the-art DLCM (described in the next section), are correct with only 70\% probability.
        
        Regarding our modified crossover operator, note that the objective of the first and second phase is to inherit correctly-reconstructed segments from the parents. We constrain the score of each of the two pieces in question to be at least 0.8, as a good starting threshold. (Note that the score is in the range between 0 to 1, due to the normalization of the compatibility measure as explained in Sec.~\ref{sec:cm_post_processing}.) Furthermore, since the algorithm improves as the number of generations goes up, ({\ie} chromosome fitness increases), the resulting threshold becomes greater than 0.8. The idea behind the dynamic threshold, is to overcome errors made in previous generations.
        Phase III is carried out if the two chromosomes agree on the same pair of pieces, {\ie} they are likely to be correct, with high probability.
        
        In the first three phases the crossover inherits adjacent pieces from the parents; however, these phases might not necessarily result in a successful addition of a new piece to the kernel. Thus, Phases IV and V, which rely solely on our proposed DLCM, could be used alternatively by considering the most compatible and second-most compatible pieces.
        
        If Phases IV and V still fail to add one more piece to the kernel (because the pieces considered are already placed in the kernel), Phase VI is invoked to complete the puzzle configuration, by placing randomly a free piece at an open boundary.

\section{Training a Compatibility Measure}
\label{sec:dl_cm}
    We have striven to develop a DL model for learning automatically a compatibility measure, such that given two puzzle pieces, it would distinguish between adjacent and non-adjacent pieces. The proposed method is based loosely on ideas from the field of {\em metric embedding learning}. The goal of metric embedding learning is to learn a function $ f_\theta(x): \mathbb{R}^F \to \mathbb{R}^D $, which maps semantically similar points from the data manifold $ \mathbb{R}^F $ onto metrically close points in $ \mathbb{R}^D $. This approach was first presented by Weinberger and Saul~\cite{weinberger2009distance}, in the context of nearest-neighbor classification. Schroff {\etal}~\cite{schroff2015facenet} subsequently proposed using a deep {\em convolutional neural network} (CNN)-based embedding of human faces, which is trained via a so-called {\em triplet-loss} described below.
    
    We propose to formulate the problem of learning a compatibility measure as learning a single-dimensional embedding $ \mathbb{E} \times \mathbb{E} \to \mathbb{R} $, where $ \mathbb{E} $ is the group of all puzzle piece edges. Here we want to ensure that given a piece-edge $ e_i $ {\em (anchor)} and its adjacent piece-edge $ e_j $ in the original image, the score of the {\em positive pair} $ (e_i,e_j) $ will be higher than any  {\em negative pair} $ (e_i,e_k) $. This can be achieved by minimizing the loss
    
    \begin{equation*}
    L = \sum_{e_i,e_j,e_k \in T} max(0, 1 - f(e_i,e_j) + f(e_i,e_k)),
    \end{equation*}
    where $ f $ is a deep convolutional neural network and $ T $ is the training set.
    
    \subsection{Triplet Selection}
        Since the number of possible triplets in the training set is quite large, we generated the training triplets online. Specifically, we selected, for each epoch, 25 pieces at random from every puzzle in our dataset.
        We used the edges of each piece as anchors, generating positive pairs from each edge and its neighboring pieces. (Usually this results in four pairs, but could also result in three or two pairs only, for pieces along the puzzle boundaries and the four corner pieces, respectively.)
        For each such positive pair, we randomly select a non-adjacent piece edge and create its accompanying negative pair to form a triplet.
        
\begin{figure}[h]
    \centering
        \begin{subfigure}[t]{0.32\linewidth}
            \includegraphics[width=\linewidth]{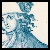}
        \end{subfigure}
        \begin{subfigure}[t]{0.32\linewidth}
            \includegraphics[width=\linewidth]{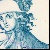}
        \end{subfigure}
        \begin{subfigure}[t]{0.32\linewidth}
            \includegraphics[width=\linewidth]{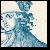}
        \end{subfigure}
    \caption{Illustration of tile augmentation via degradation and shifting. From left to right: Degraded tile by removing 2 pixels from its outer frame; shifted tile by one pixel to the left, and one pixel up; augmented tile with degradation and shifting.}
    \label{fig:augmented_tiles}
    \end{figure}

        Next, we randomly augmented each piece in each pair, using either {\em degradation} or {\em shifting}. Degradation replaces randomly the outermost pixel frame of the piece with zeros. With uniform probability, we may replace no pixel, replace a pixel-wide frame, or replace a double pixel frame. This should aid the network in learning more than only near-border textures. For shifting, we randomly shift the piece anywhere between zero to two pixels horizontally or vertically (filling with zeros empty locations). Figure~\ref{fig:augmented_tiles} demonstrates some possible outcomes.
    
    \subsection{Deep Convolutional Neural Networks}
        We trained a deep {\em convolutional neural network} (CNN), which receives as input a pair of puzzle pieces and returns a real number score. All pieces are of size $ 50 \times 50 $ pixels. Although most actual puzzle pieces are larger, we downscaled them to better fit in memory and speed up the training phase. Always taking the anchor piece to be on the left, we rotate the pieces accordingly. For example, to compare the left edge of an anchor piece with the right edge of piece $p$, we would rotate both pieces by 180\degree, so that the anchor piece will still be on the left, but its left edge now points to the right.
        
\begin{figure}[h]
        \centering
        \includegraphics[width=0.45\textwidth]{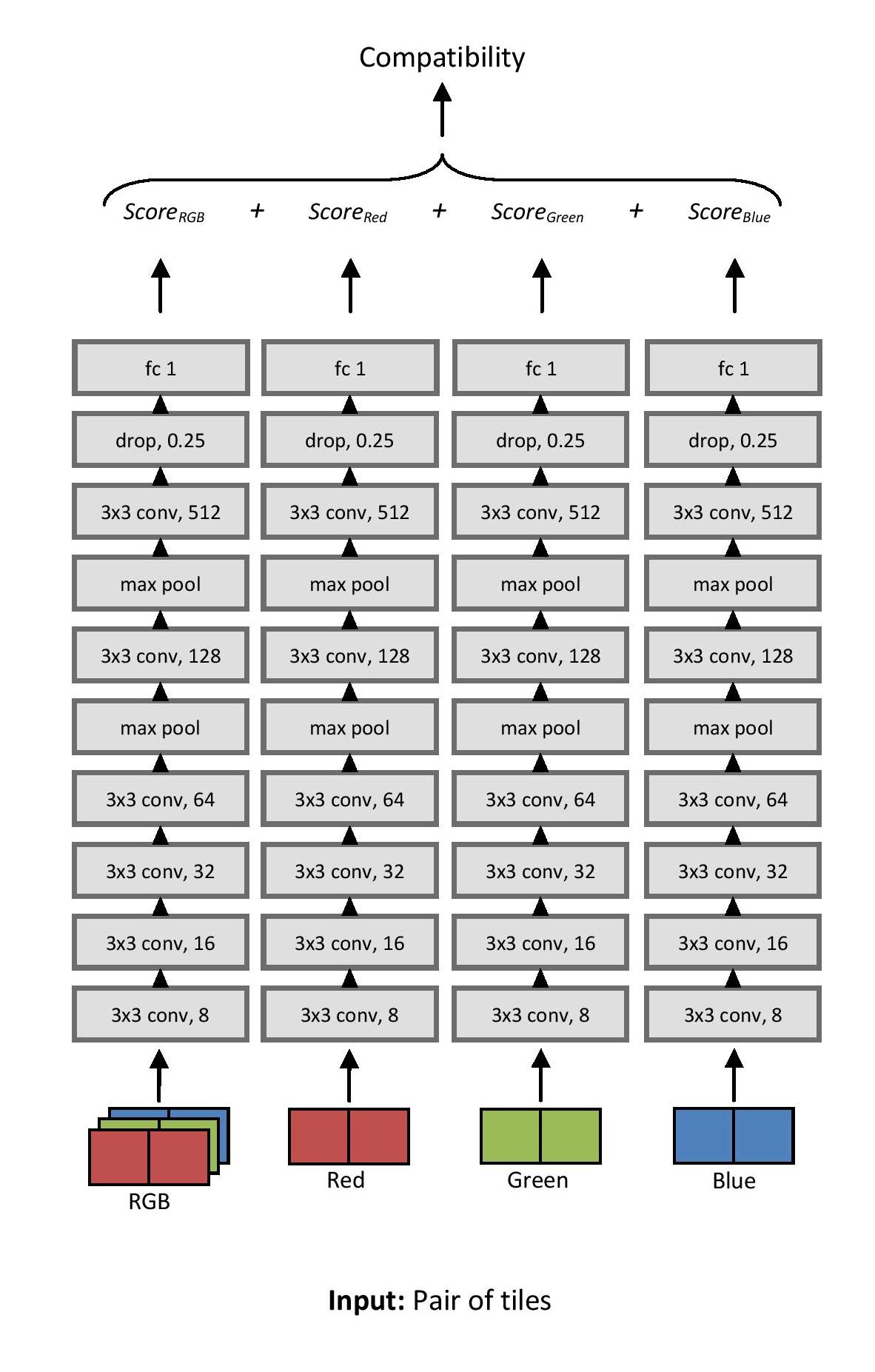}
        \caption{DLCM architecture. The input is a pair of two squared tiles of size $50 \times 50$ pixels ({\ie} input dimension is $50 \times 100 \times 3$). The DLCM network contains 3.4M parameters, and uses non-linear ReLU activation functions with no bias.}
        \label{fig:CNN_configuration}
        \end{figure}

        During training we noticed that determining the degree of compatibility for some pairs could be rather difficult for both the network and human experts, but it becomes quite easier for humans when looking only at a single color channel. Drawing on this observation, we trained the following networks: Red-Net, Green-Net, and Blue-Net (named after the color channels each receives as input), as well as a fourth network, RGB-Net (which receives all three channels as input). All networks share the exact same architecture, as depicted in Figure~\ref{fig:CNN_configuration}. During training we presented all networks with the same batch ({\ie} same training samples); each network's loss was calculated separately, so as not to affect the other networks.
        Table~\ref{tab:cm_compare} gives a performance comparison, regarding the above individual networks and their proposed combined scheme.

        We trained all networks using {\em stochastic gradient descent} (SGD) with standard backpropagation~\cite{rumelhart1986learning} and Adam~\cite{DBLP:journals/corr/KingmaB14}, using a learning rate of 0.0001. We used a batch size of 64 and ran for a total of 850 epochs. For training we used a modern PC with 3.5GHz CPU, 32GB RAM, and a single GPU with 11GB memory.
    
    \subsection{Post-processing}
    \label{sec:cm_post_processing}
        To enhance the global optimization, we first apply a per-edge normalization of all compatibility scores to the range between 0 and 1, using the min-max normalization. Namely, for each piece edge $e_i$ we calculate its compatibility with every other edge, extract the minimum and maximum across all compatibility scores, and normalize according to
    
        \begin{equation*}
        C'(e_i,e_j) = \dfrac{C(e_i,e_j) - min(C(e_i,*))}{max(C(e_i,*)) - min(C(e_i,*))}.
        \end{equation*}
        
        Next, we note that the framework described above offers no symmetry guarantee, {\ie} that for any two piece edges $e_i$ and $e_j$, $C(e_i,e_j) = C(e_j,e_i)$. Assuming that any deviation in symmetry is mostly erroneous, we manually enforce symmetry by averaging the two scores and defining the following symmetric compatibility measure
        
        \begin{equation*}
        C''(e_i,e_j) = C''(e_j,e_i) = \dfrac{C'(e_i,e_j) + C'(e_i,e_j)}{2}.
        \end{equation*}

\section{Datasets} \label{sec:datasets}
    We acquired eight high-resolution images from the National Tile Museum (Museu Nacional do Azulejo, MNAz), Lisbon, Portugal, which were kept as test data for the final reconstruction, {\ie} they were not used during the CNN training. The size of each image is given in Table~\ref{tab:test}.

    \begin{table}[]
    \centering
    \begin{tabular}{|l||c|c|c|c|}
    \hline
    \multirow{2}{*}{\textbf{Image}} & \multirow{2}{*}{\textbf{Rows}} & \multirow{2}{*}{\textbf{Columns}} & \textbf{Total} & \textbf{Piece Size} \\ 
    &  &  & \textbf{Pieces} &  \textbf{(pixels)} \\ 
            \hline \hline 
    Image 0 & 8    & 18      & 144          &  650 \\ \hline
    Image 1 & 16   & 16      & 256          &  150 \\ \hline
    Image 2 & 9    & 12      & 108          &  240 \\ \hline
    Image 3 & 11   & 29      & 319          &  100 \\ \hline
    Image 4 & 15   & 10      & 150          &  165 \\ \hline
    Image 5 & 9    & 23      & 207          &  225 \\ \hline
    Image 6 & 9    & 18      & 162          &  280 \\ \hline
    Image 7 & 12   & 10      & 120          &  240 \\ \hline
    \end{tabular}
    \caption{Image details of test set received from the MNAz.}
    \label{tab:test} 
    \end{table}

    We acquired nine additional images of smaller size from the MNAz: Five images of 25 pieces each and four images of 40, 48, 60, and 72 pieces, respectively. Due to the relatively small number of pieces per image, these images might not be adequately representative of the actual reconstruction problem. Nevertheless, given their acquisition from the museum, we regard them as sufficiently representative, in terms of content, and thus use them as a held-out validation set during the CNN training.
    
\begin{figure}
    \centering
        \begin{subfigure}[t]{0.59\linewidth}
            \includegraphics[width=\linewidth]{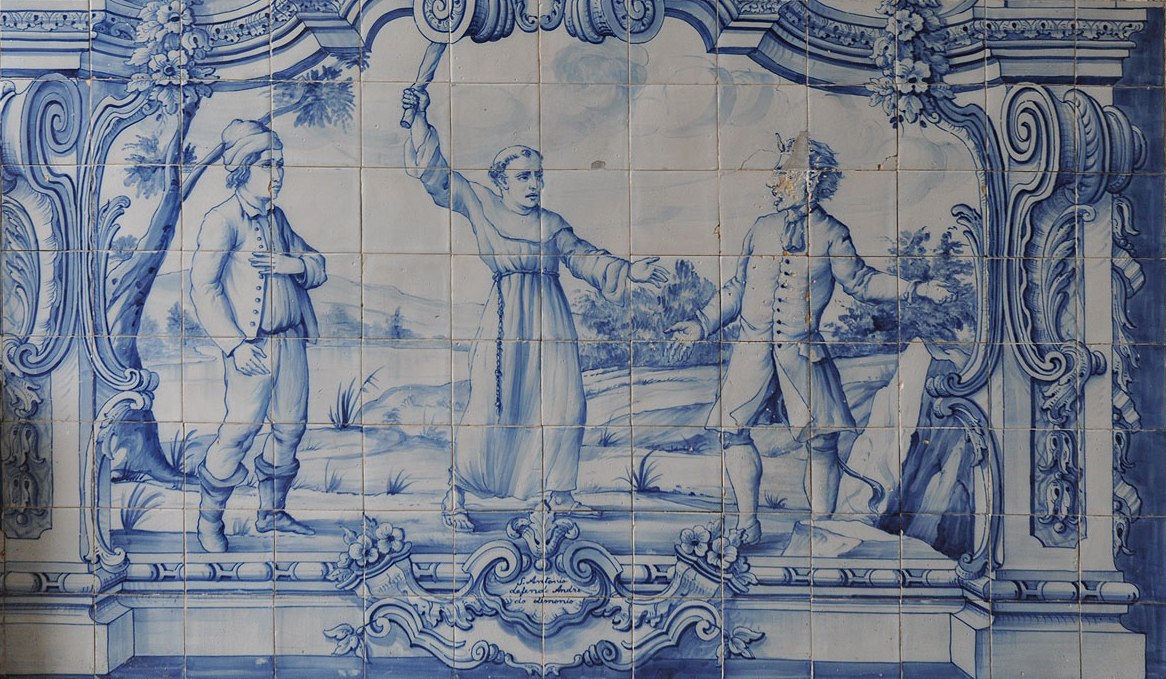}
        \end{subfigure}
        \begin{subfigure}[t]{0.39\linewidth}
            \includegraphics[width=\linewidth]{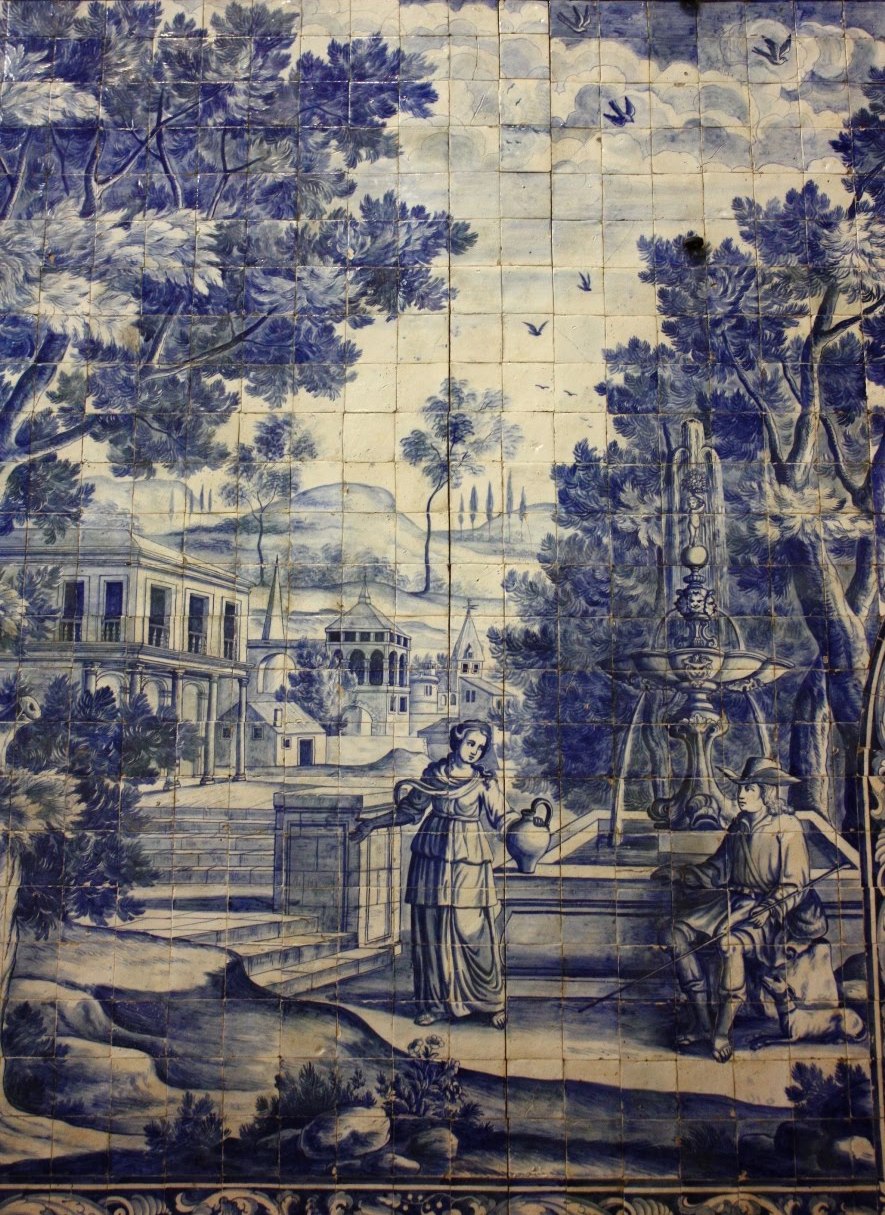}
        \end{subfigure}
        \begin{subfigure}[t]{0.98\linewidth}
            \includegraphics[width=\linewidth]{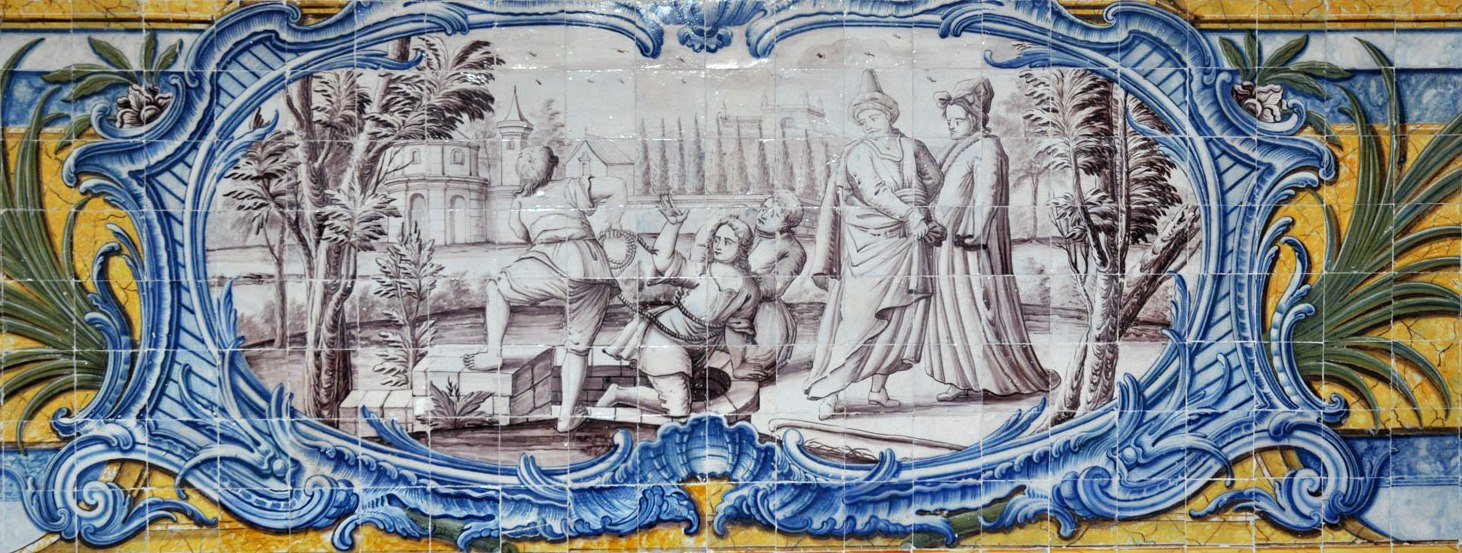}
        \end{subfigure}    
    \caption{Training set images downloaded from the Internet.}
    \label{fig:train_images}
    \end{figure}

    In addition to the above datasets, we also downloaded 89 images of Portuguese tile panels from the Internet, some of which were photographed by tourists. Figure~\ref{fig:train_images} depicts a few downloaded images. We manually went through each puzzle, and counted the number of pieces per row and column. Knowing also the image dimensions, we could easily resize each image to $50 \times 50$ pixels. We picked nine of these images, ``cut'' them manually to pieces along tile lines, and added the resulting images to the validation set. The other 80 images were used as a training set for the CNN; cutting automatically the images, we gathered a total of 9,031 pieces. The automatic cutting may not always overlap fully with the actual piece boundaries, but this does not occur too often and might even aid in avoiding overfitting.
    
    To summarize, for the training of the compatibility measure, we used a training set of 80 images and a validation set of 18 images (nine from the MNAz and nine from the Internet). For the evaluation of the compatibility measure and the overall solver's reconstruction capability, we use a test set of the eight high-resolution images acquired from the MNAz.

    \begin{figure}
    \centering
        \begin{subfigure}[t]{0.59\linewidth}
            \includegraphics[width=\linewidth]{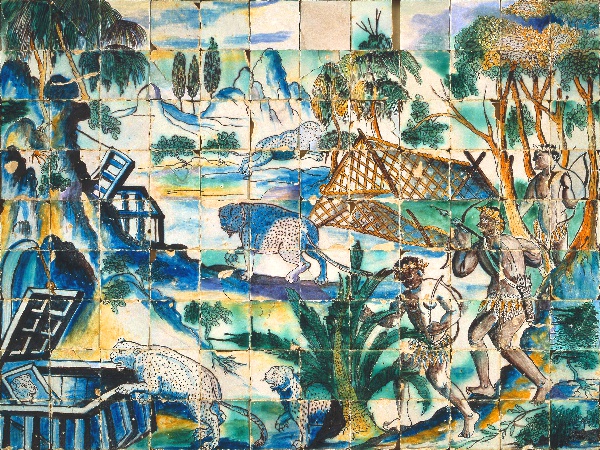}
        \end{subfigure}    
        \begin{subfigure}[t]{0.39\linewidth}
            \includegraphics[width=\linewidth]{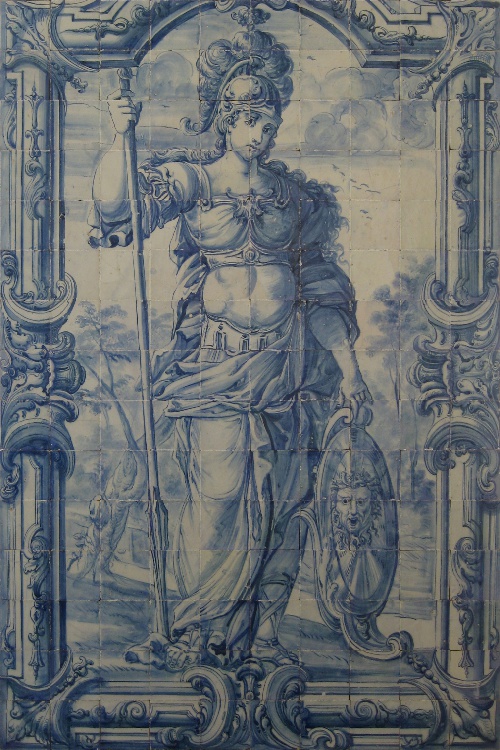}
        \end{subfigure}    
        \begin{subfigure}[t]{0.98\linewidth}
            \includegraphics[width=\linewidth]{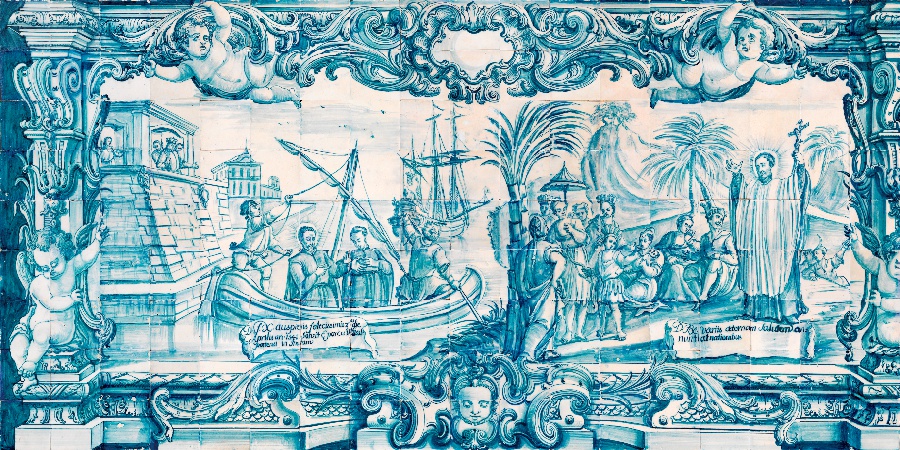}
        \end{subfigure}    
    \caption{Test set images received from the MNAz.}
    \label{fig:test_images}
    \end{figure}

\section{Experimental Results}
\label{sec:experimental_results}
    \subsection{Compatibility Measure Evaluation}
    \label{sec:experimental_results_cm}
        Previous works~\cite{conf/cvpr/PomeranzSB11} evaluate compatibility measures by their accuracy. For each piece edge, we rank all other piece edges according to the measure in question, and report the frequency of occurrence (in percentage) that the piece edge ranked as the most compatible was indeed the correct edge. We used a generalized metric, which we call {\em $rank_\alpha$ score}, to report the percentage of actual neighboring edges found at each location of the sorted array. In other words, we define $rank_i$ of a measure as the ground truth fraction of adjacent edges which were ranked $i$-th most compatible according to the measure. Thus, the standard accuracy criterion for a given measure would be $rank_1$, since a perfect measure should have $rank_1 = 100\%$ and $rank_i = 0\%$ for all $i > 1$.
    
        \begin{figure*}
        \centering
            \begin{subfigure}[t]{0.30\linewidth}
                \includegraphics[width=\linewidth]{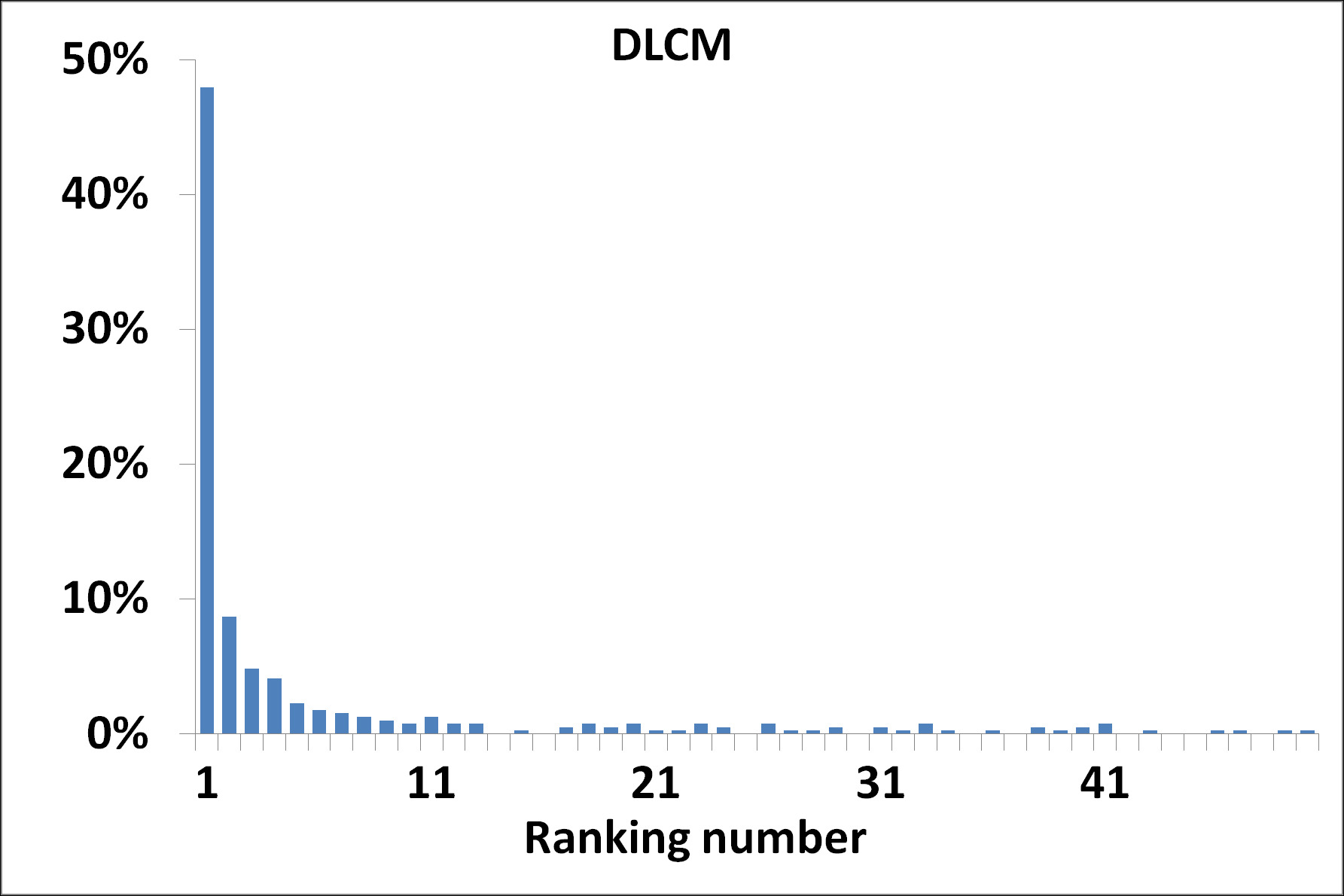}
            \end{subfigure} 
            ~
            \begin{subfigure}[t]{0.30\linewidth}
                \includegraphics[width=\linewidth]{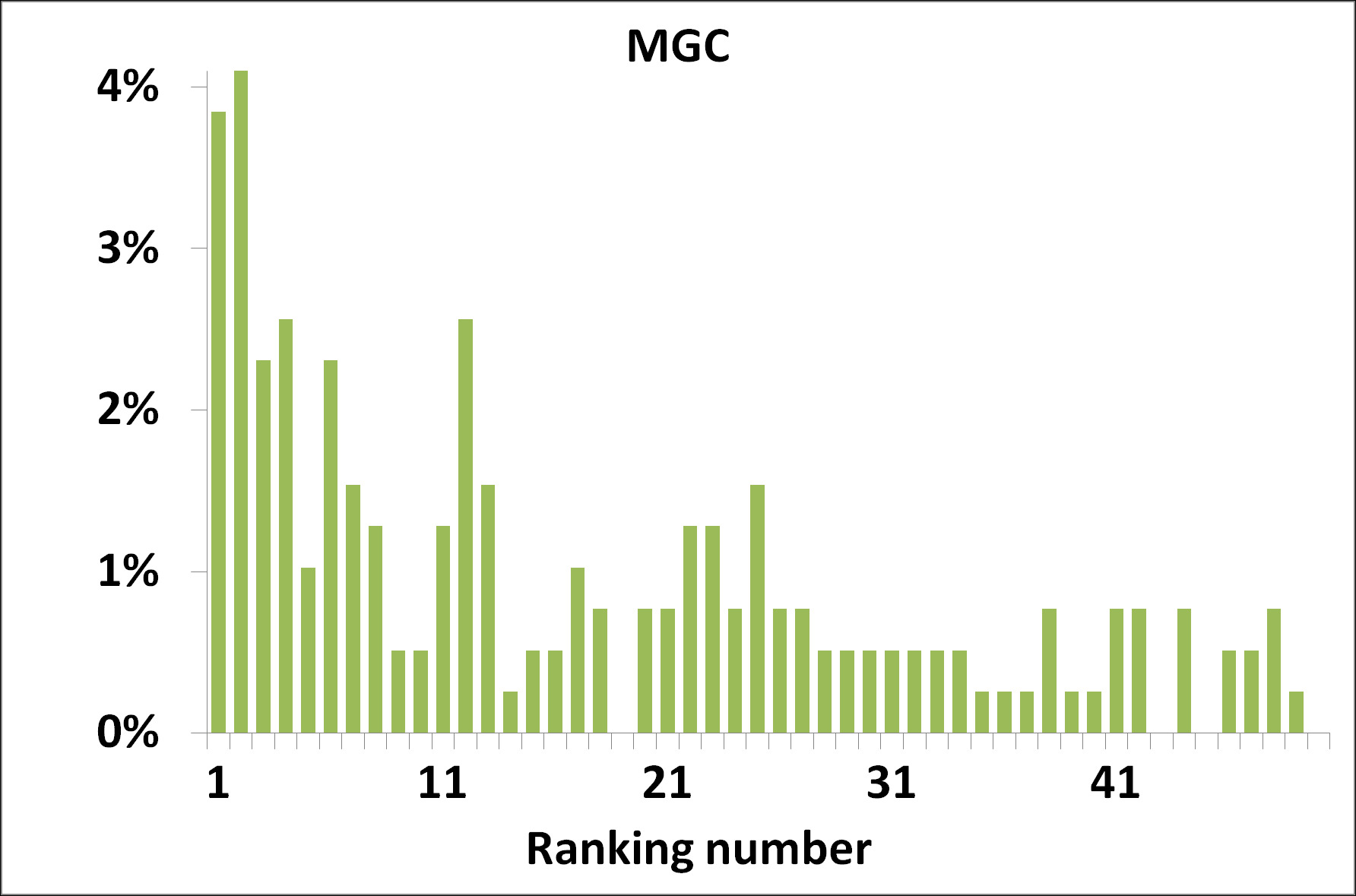}
            \end{subfigure} 
            ~
            \begin{subfigure}[t]{0.30\linewidth}
                \includegraphics[width=\linewidth]{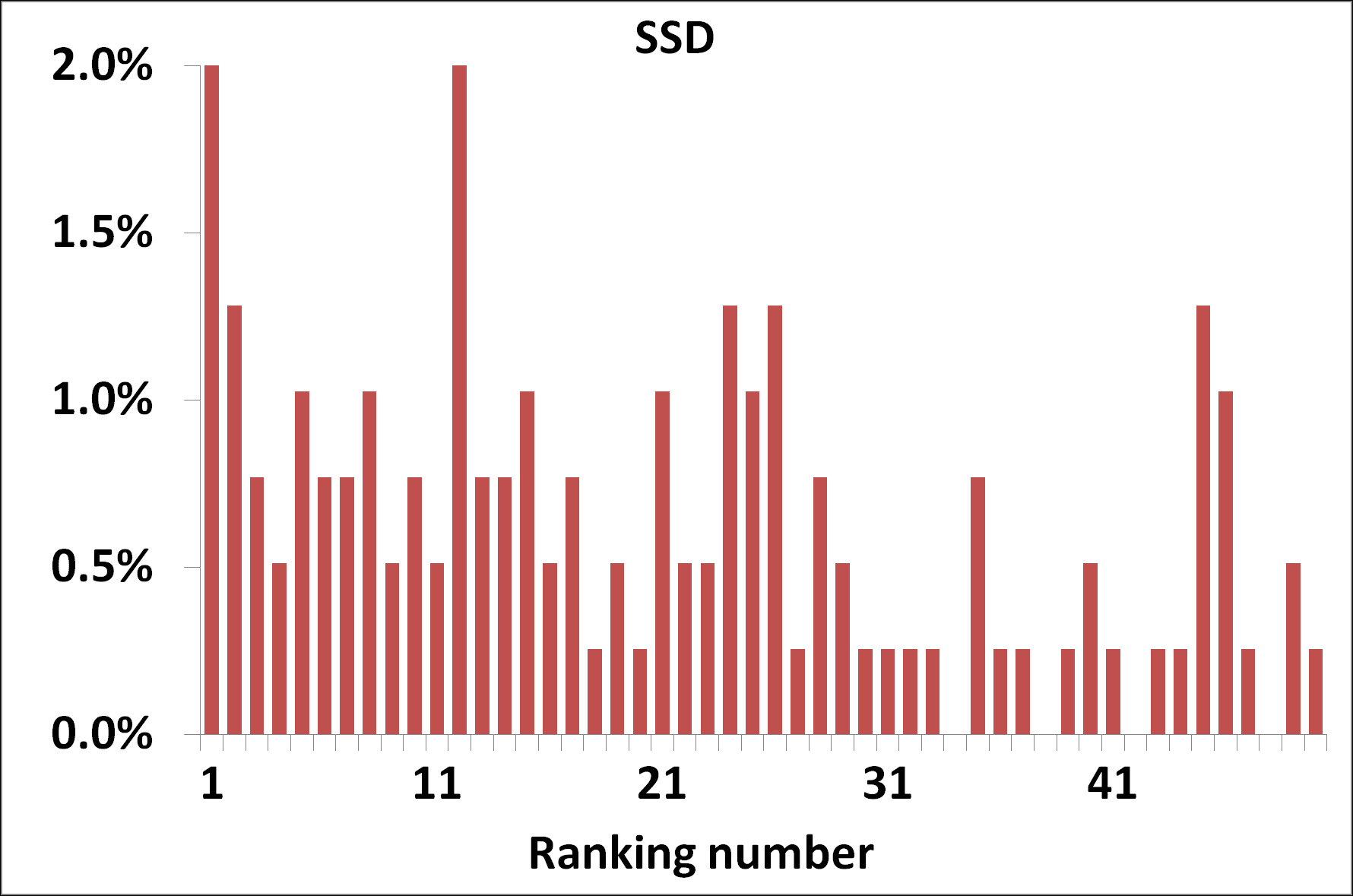}
            \end{subfigure} 
            
            ~
            \begin{subfigure}[t]{0.91\linewidth}
                \includegraphics[width=\linewidth]{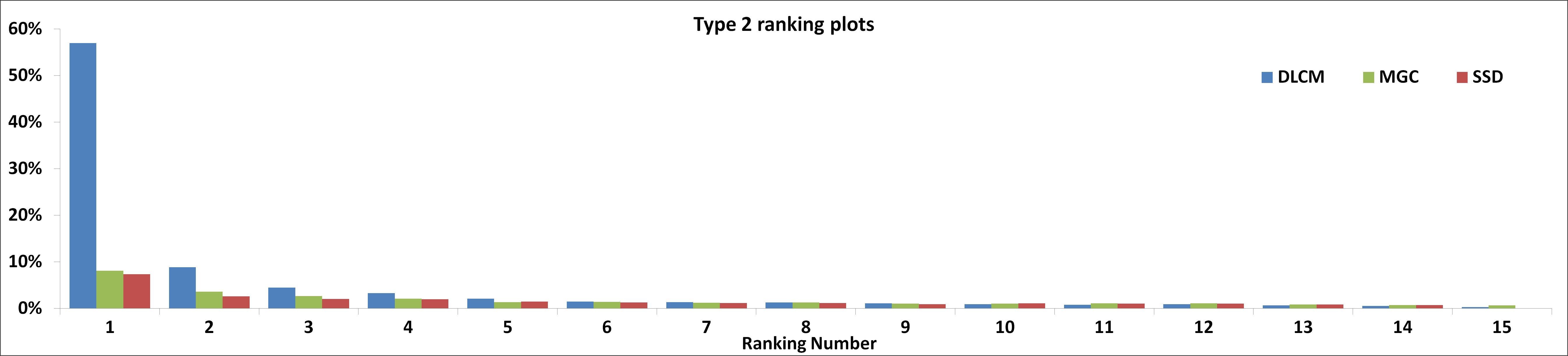}
            \end{subfigure} 
    
        \caption{Rank percentages using our DLCM vs. SSD and the MGC measures for Type 2 puzzles. Top three plots correspond to a single test image (with unknown piece orientation). Bottom plot corresponds to average ranking percentage over all eight test images (with unknown piece orientation). Note the clear-cut superior performance of DLCM. Interestingly,  $rank_2$ percentage of our CNN model is greater than the $rank_1$ percentage obtained for the SSD and MGC measures.}
        \label{fig:rank_alpha_scores}
        \end{figure*}
    
        We trained our CNN-based compatibility measure as previously described, and evaluated it on our test set images, which were not used at all during the training phase. Our compatibility measure achieves $rank_1$ of 68.45\%, assuming known piece orientation (Type 1 variant) and $rank_1$ of 56.9\%, relaxing this assumption (Type 2 variant). We compared our results to the SSD~\cite{conf/cvpr/PomeranzSB11} measure, which achieves 12.7\% and 7.3\%, respectively, and the MGC measure~\cite{conf/cvpr/Gallagher12}, which achieves 17.4\% and 9.1\%, respectively. Also, we compared between the performances of the individual sub-networks of our CNN model. The entire comparison is summarized in Table~\ref{tab:cm_compare}.

        \begin{table}
        \centering
        \begin{tabular}{|c||c|c|}
        \hline
        & \textbf{Type 1} & \textbf{Type 2} \\ \hline \hline
        SSD~\cite{conf/cvpr/PomeranzSB11} & 12.7\% & 7.3\% \\ \hline
        MGC~\cite{conf/cvpr/Gallagher12} & 17.4\% & 9.1\% \\ \hline \hline
        Red-Net & 56.9\% & 44.1\% \\ \hline
        Green-Net & 57.2\% & 45.1\% \\ \hline
        Blue-Net & 53.4\% & 40.8\% \\ \hline
        RGB-Net & 59.5\% & 47.5\% \\ \hline
        \textbf{DLCM} & \textbf{68.4\%} & \textbf{56.9\%} \\ \hline
        \end{tabular}
        \caption{Comparison of $rank_1$ scores of our DLCM with those for the SSD and MGC measures; also included are $rank_1$ scores of the DLCM's four sub-networks ({\ie} Red-Net, Green-Net, Blue-Net, and RGB-Net), demonstrating the added value of their combination.}
        \label{tab:cm_compare}
        \vspace{-5pt}
        \end{table}

        Next, we compared the different $rank_\alpha$ scores of our measure versus those obtained for SSD and MGC. Figure~\ref{fig:rank_alpha_scores} presents these scores for a single test image and the average of these scores over the entire test set. The plots obtained attest to the relatively high quality of the learned measure, having the highest $rank_1$ score and monotonically-decreasing lower ranks, unlike the more uniform distribution obtained for the other measures.
        
        Also, to verify the assumption that led to the post-processing steps described in Section~\ref{sec:cm_post_processing}, we evaluated the raw measure obtained by the CNN. The values obtained for this measure were 62.8\% and 50.6\%, respectively, for the Type 1 and Type 2 problem variants. These results strongly support the use of the post-processing step, according to Subsection~\ref{sec:cm_post_processing}.
        
        The results clearly indicate that our trained measure is by far superior to other established compatibility measures, both quantitatively, in terms of higher accuracy, as well as qualitatively in terms of a smoother distribution.
    
    \subsection{Puzzle Reconstruction}
        We incorporated our newly trained compatibility measure into our enhanced GA framework, in an attempt to reconstruct each of the test set images. We report the reconstruction accuracy, according to the {\em neighbor comparison} definition applied in previous works, namely the fraction of correctly assigned neighbors, {\ie} the fraction of ground truth adjacent edges in our solution.
        
        We attempted reconstruction under four different variants of the problem. In all variants we assumed an unknown location of the different pieces. The variants differ with respect to a priori knowledge of piece orientation and puzzle dimensions. Obviously, the hardest variant, which is most reflective of a real-world scenario, is the one for which both piece orientation and puzzle dimensions are unknown.
        
        We ran our GA version ten times on each image, and reported the best result. For comparison, we also tried reconstructing the images using the solver proposed by Gallagher~\cite{conf/cvpr/Gallagher12}. We chose to compare against this solver, because it is one of the few solvers that supports all of the different variants and whose reported performance is still competitive relatively to state-of-the-art on available JPP benchmarks and the Portuguese tile panels in~\cite{andalo2016impa}. To justify the net added value of our proposed {\em kernel-growth} GA solver, we compared also its performance (using our DLCM) with that of the GA solvers~\cite{Sholomon_2013_CVPR, sholomon2014generalized, sholomon2014genetic}.
        The comparative results for all four cases are reported in Table~\ref{tab:reconstruction_results}. Examples of  reconstructed panels are shown in Figure~\ref{fig:ga_reconstruction_samples}.

        \begin{table*}
        \centering
        \begin{tabular}{|c||c|c||c|c|}
        \hline
          \multirow{3}{*}{Method}
        & \multicolumn{2}{|c||}{Type 1} & \multicolumn{2}{|c|}{Type 2} \\
         \cline{2-5}
         & Known & Unknown & Known & Unknown \\
         & dims.  & dims.  & dims.  & dims. \\ \hline \hline
        
        Gallagher+ & \multirow{2}{*}{---} & \multirow{2}{*}{13.0\%}  & \multirow{2}{*}{---} & \multirow{2}{*}{3.5\%}  \\
        MGC        & & & & \\ \hline
        
        Kernel-growth~\cite{Sholomon_2013_CVPR, sholomon2014generalized}+ &\multirow{2}{*}{84.5\%} & \multirow{2}{*}{---} & \multirow{2}{*}{58.6\%} & \multirow{2}{*}{---} \\
        symmetric DLCM & & & & \\ \hline
        
        Multi-segment~\cite{sholomon2014genetic}+ & \multirow{2}{*}{---} & \multirow{2}{*}{---} & \multirow{2}{*}{---} & \multirow{2}{*}{62.9\%} \\
        symmetric DLCM & & & & \\ \hline
        
        Our kernel-growth+ & \multirow{2}{*}{\textbf{96.9\%}} & \multirow{2}{*}{\textbf{96.2\%}} & \multirow{2}{*}{66.5\%} & \multirow{2}{*}{70.6\%} \\
        DLCM & & & & \\ \hline
        
        Our kernel-growth+ & \multirow{2}{*}{96.3\%} & \multirow{2}{*}{96.0\%} & \multirow{2}{*}{\textbf{86.8\%}} & \multirow{2}{*}{\textbf{82.2\%}} \\
        symmetric DLCM & & & & \\ \hline
        
        \end{tabular}
        \caption{Reconstruction comparison (from top to bottom): Gallagher's greedy solver, using the MGC compatibility measure~\cite{conf/cvpr/Gallagher12}; kernel-growth GA (due to Sholomon {\etal}) with our proposed (symmetric) DLCM; multi-segment GA (due to Sholomon {\etal}) with our (symmetric) DLCM; our proposed kernel-growth GA with (non-symmetric) DLCM, and same hybrid scheme with symmetric post-processing.}
        \label{tab:reconstruction_results}
        \end{table*}

\begin{figure}[h]
        \centering
        \includegraphics[width=0.5\textwidth]{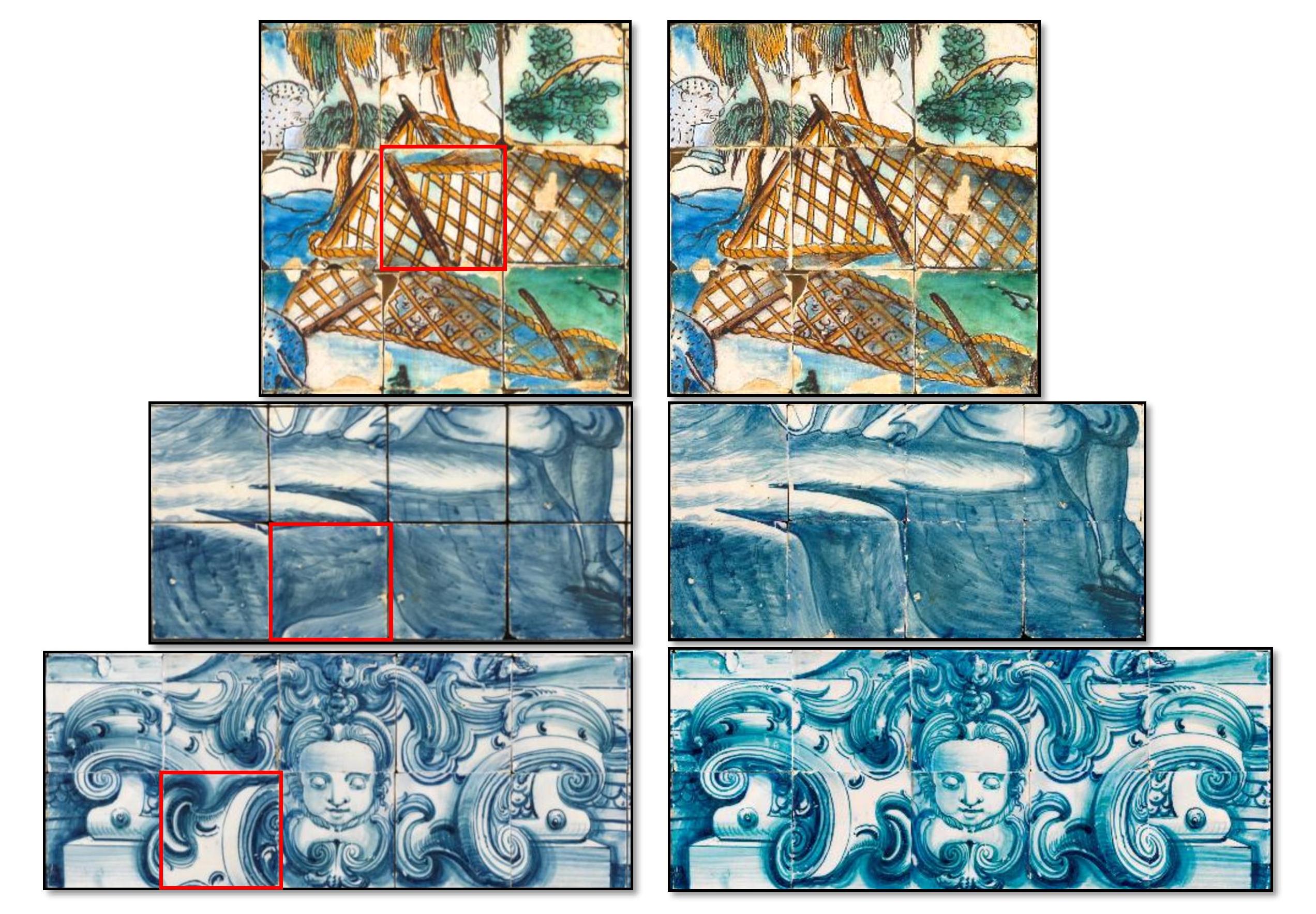}
        \caption{\textbf{Left:} Images with human errors (highlighted by red), received from the MNAz. \textbf{Right:} Correct assembly by our system for Type 2 puzzle with known dimensions.}
        \label{fig:better_reconstruction}
        \end{figure}

        Interestingly, while inspecting the reconstructed puzzles, we noticed three puzzles that were reported as not perfectly solved, despite the fact that their overall global score was greater than ground truth.
        Further manual inspection revealed that apparently, the image was not assembled correctly by the museum staff, and that the solution suggested by our algorithm was indeed the correct one. Figure~\ref{fig:better_reconstruction} shows these segments in question.
    
\section{Conclusions}
\label{sec:conclusions}
    We presented in this paper a novel hybrid scheme, based on an enhanced GA solver and a novel DL compatibility measure, for solving the challenging, real-world task of the reconstruction of Portuguese tile panels, which is a high-profile national endeavor of significant importance to Portugal's cultural heritage. Specifically, we demonstrated how to integrate successfully the above innovative components to achieve ground-breaking performance (over 96\% accuracy for Type 1 variant and roughly 87\% and 82\% accuracies, for Type 2 variant with known and unknown dimensions, respectively), for tile panels containing hundreds of relatively low-resolution tiles. Finally, we have compiled a decent benchmark of Portuguese tile panels, to be used by the Computer Vision and Evolutionary Computation communities for training and testing.
    
    With regards to future work, we intend to improve our DL-based compatibility (by considering, for example, additional training data), in an attempt to enhance the overall performance of our GA solver. In addition, we intend to extend the capabilities of our system to handle also missing tiles and mixed panels of tiles, to meet as many practical challenges as possible associated with the Portuguese tile problem.